\documentclass[11pt,a4paper]{article}
\pdfoutput=1
\usepackage[hyperref]{acl2020}
\usepackage{times}
\usepackage{latexsym}
\usepackage{amsmath}
\usepackage{amssymb}
\usepackage{pgfplots}
\usepackage{makecell}
\usepackage{threeparttable}
\usepackage[caption=false]{subfig}
\usepackage{scalefnt}
\usepackage{boldline}
\usepackage{xcolor}
\usepackage{booktabs}
\usepackage{graphicx}
\usepackage{mathtools}
\graphicspath{ {./images/} }
\usepackage[ruled,vlined]{algorithm2e}
\pgfplotsset{width=7.5cm,compat=1.9}

\newcommand{\floor}[1]{\lfloor #1 \rfloor}
\definecolor{mycolor1}{RGB}{27,158,119}
\definecolor{mycolor2}{RGB}{217,95,2}
\definecolor{mycolor3}{RGB}{117,112,179}
\definecolor{mycolor5}{RGB}{228,26,28}
\definecolor{mycolor4}{RGB}{55,126,184}
\definecolor{mycolor6}{RGB}{77,175,74}
\definecolor{mycolor7}{RGB}{152,78,163}
\usepackage{microtype}
\usepackage{multirow}

\aclfinalcopy 
\def\aclpaperid{1641} 

\newcommand\blfootnote[1]{%
  \begingroup
  \renewcommand\thefootnote{}\footnote{#1}%
  \addtocounter{footnote}{-1}%
  \endgroup
}

\newcommand{\ours}{\textsc{BioSyn}}
\newcommand{\lowerb}{0.8}
\newcommand{\upperb}{2.6}

\newcommand{\accone}{\text{Acc@1}}
\newcommand{\accfive}{\text{Acc@5}}

\title{Biomedical Entity Representations with Synonym Marginalization}

\author{Mujeen Sung \quad Hwisang Jeon \quad Jinhyuk Lee$^\dagger$ \quad Jaewoo Kang$^\dagger$\\
Korea University\\
\texttt{\{mujeensung,j\_hs,jinhyuk\_lee,kangj\}@korea.ac.kr}
}

\date{}

\begin{document}
\maketitle
\begin{abstract}
Biomedical named entities often play important roles in many biomedical text mining tools.
However, due to the incompleteness of provided synonyms and numerous variations in their surface forms, normalization of biomedical entities is very challenging.
In this paper, we focus on learning representations of biomedical entities \emph{solely based on the synonyms of entities}.
To learn from the incomplete synonyms, we use a model-based candidate selection and maximize the marginal likelihood of the synonyms present in top candidates.
Our model-based candidates are iteratively updated to contain more difficult negative samples as our model evolves.
In this way, we avoid the explicit pre-selection of negative samples from more than 400K candidates.
On four biomedical entity normalization datasets having three different entity types (disease, chemical, adverse reaction), our model \ours~consistently outperforms previous state-of-the-art models almost reaching the upper bound on each dataset.
\end{abstract}
\blfootnote{\textsuperscript{$\dagger$}Corresponding authors}

\section{Introduction}
Biomedical named entities are frequently used as key features in biomedical text mining.
From biomedical relation extraction~\citep{xu2016cd,li2017neural} to literature search engines~\citep{lee2016best}, many studies are utilizing biomedical named entities as a basic building block of their methodologies.
While the extraction of the biomedical named entities is studied extensively~\citep{sahu-anand-2016-recurrent,habibi2017deep}, the normalization of extracted named entities is also crucial for improving the precision of downstream tasks~\citep{leaman2013dnorm,wei2015gnormplus}.

Unlike named entities from general domain text, typical biomedical entities have several different surface forms, making the normalization of biomedical entities very challenging.
For instance, while two chemical entities `\textit{motrin}' and `\textit{ibuprofen}' belong to the same concept ID (MeSH:D007052), they have completely different surface forms.
On the other hand, mentions having similar surface forms could also have different meanings (e.g. `\textit{dystrophinopathy}' (MeSH:D009136) and `\textit{bestrophinopathy}' (MeSH:C567518)).
These examples show a strong need for building latent representations of biomedical entities that capture semantic information of the mentions.

In this paper, we propose a novel framework for learning biomedical entity representations based on the synonyms of entities.
Previous works on entity normalization mostly train binary classifiers that decide whether the two input entities are the same (positive) or different (negative)~\citep{leaman2013dnorm,li2017cnn,fakhraei2019nseen,phan2019robust}.
Our framework called \ours~uses the synonym marginalization technique, which maximizes the probability of all synonym representations in top candidates.
We represent each biomedical entity using both sparse and dense representations to capture morphological and semantic information, respectively.
The candidates are iteratively updated based on our model's representations removing the need for an explicit negative sampling from a large number of candidates.
Also, the model-based candidates help our model learn from more difficult negative samples.
Through extensive experiments on four biomedical entity normalization datasets, we show that \ours~achieves new state-of-the-art performance on all datasets, outperforming previous models by \lowerb\%$\sim$\upperb\% top1 accuracy.
Further analysis shows that our model's performance has almost reached the performance upper bound of each dataset.

The contributions of our paper are as follows:
First, we introduce \ours~for biomedical entity representation learning, which uses synonym marginalization dispensing with the explicit needs of negative training pairs.
Second, we show that the iterative candidate selection based on our model's representations is crucial for improving the performance together with synonym marginalization.
Finally, our model outperforms strong state-of-the-art models up to \upperb\% on four biomedical normalization datasets.\footnote{Code available at \href{https://github.com/dmis-lab/BioSyn}{https://github.com/dmis-lab/BioSyn}.}

%

\begin{figure*}[!t]
\centering
\includegraphics[width=13cm,trim=0 0 0 20]{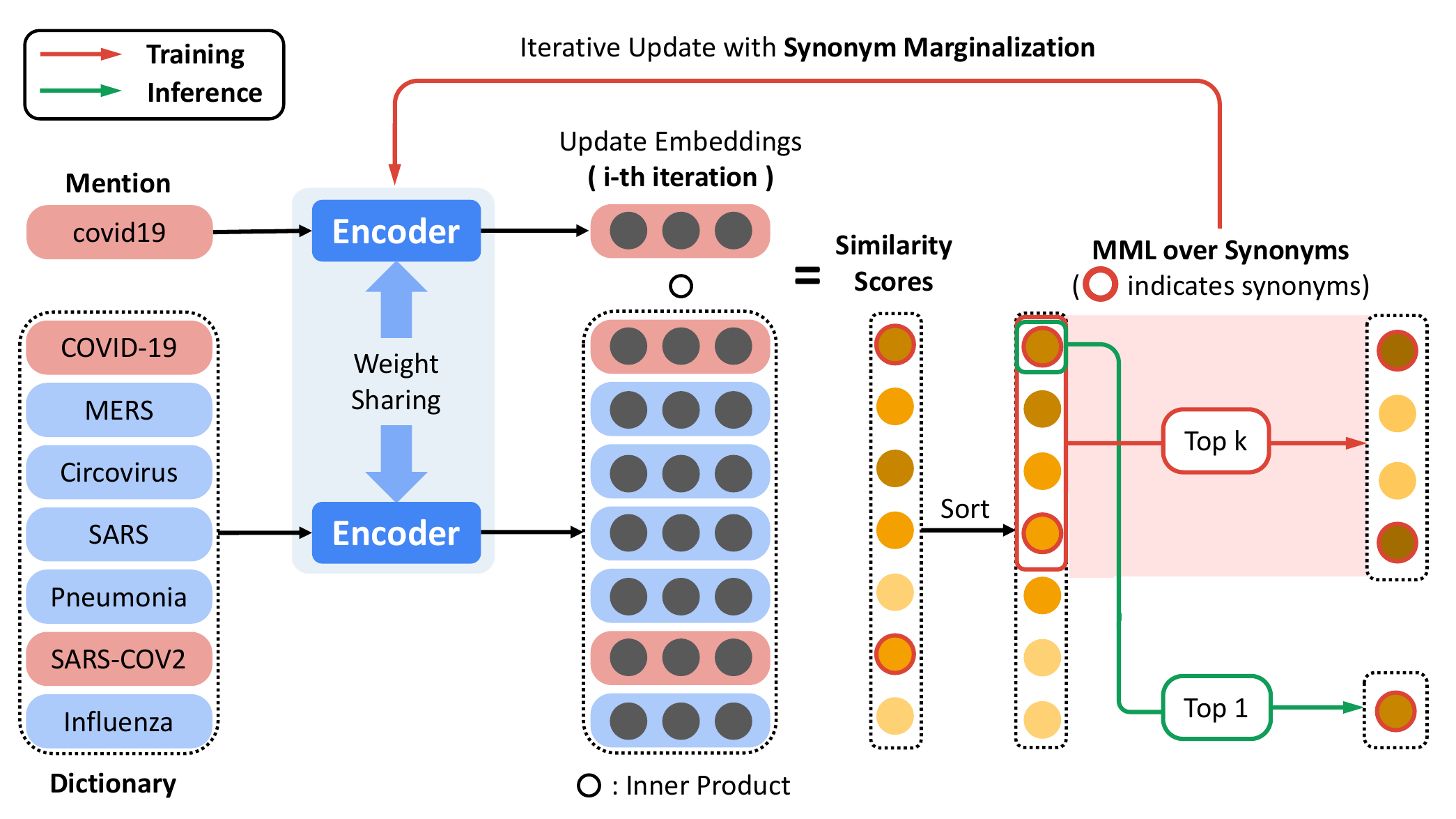}
\caption{The overview of \ours. An input mention and all synonyms in a dictionary are embedded by a shared encoder and the nearest synonym is retrieved by the inner-product. Top candidates used for training are iteratively updated as we train our encoders.
}
\label{fig:biosyn}
\end{figure*}

\section{Related Works}

Biomedical entity representations have largely relied on biomedical word representations.
Right after the introduction of Word2vec~\citep{mikolov2013distributed}, \citet{pyysalo2013distributional} trained Word2Vec on biomedical corpora such as PubMed.
Their biomedical version of Word2Vec has been widely used for various biomedical natural language processing tasks~\citep{habibi2017deep,wang2018cross,giorgi2018transfer,li2017neural} including the biomedical normalization task~\citep{mondal2019medical}.
Most recently, BioBERT~\citep{10.1093/bioinformatics/btz682} has been introduced for contextualized biomedical word representations.
BioBERT is pre-trained on biomedical corpora using BERT~\citep{devlin2018bert} and numerous studies are utilizing BioBERT for building state-of-the-art biomedical NLP models~\citep{lin2019bert,jin2019probing,alsentzer2019publicly,sousa2019silver}.
Our model also uses pre-trained BioBERT for learning biomedical entity representations.

The intrinsic evaluation of the quality of biomedical entity representations is often verified by the biomedical entity normalization task~\citep{leaman2013dnorm, phan2019robust}.
The goal of the biomedical entity normalization task is to map an input mention from a biomedical text to its associated CUI (Concept Unique ID) in a dictionary.
The task is also referred to as the entity linking or the entity grounding~\citep{dsouza-ng-2015-sieve,leaman2016taggerone}.
However, the normalization of biomedical entities is more challenging than the normalization of general domain entities due to a large number of synonyms.
Also, the variations of synonyms depend on their entity types, which makes building type-agnostic normalization model difficult~\citep{leaman2013dnorm,li2017cnn,mondal2019medical}.
Our work is generally applicable to any type of entity and evaluated on four datasets having three different biomedical entity types.

While traditional biomedical entity normalization models are based on hand-crafted rules~\citep{dsouza-ng-2015-sieve, leaman2015tmchem}, recent approaches for the biomedical entity normalization have been significantly improved with various machine learning techniques.
DNorm~\citep{leaman2013dnorm} is one of the first machine learning-based entity normalization models, which learns pair-wise similarity using tf-idf vectors.
Another machine learning-based study is CNN-based ranking method~\citep{li2017cnn}, which learns entity representations using a convolutional neural network.
The most similar works to ours are NSEEN~\citep{fakhraei2019nseen} and BNE~\citep{phan2019robust}, which map mentions and concept names in dictionaries to a latent space using LSTM models and refines the embedding using the negative sampling technique.
However, most previous works adopt a pair-wise training procedure that explicitly requires making negative pairs.
Our work is based on marginalizing positive samples (i.e., synonyms) from iteratively updated candidates and avoids the problem of choosing a single negative sample.

In our framework, we represent each entity with sparse and dense vectors which is largely motivated by techniques used in information retrieval.
Models in information retrieval often utilize both sparse and dense representations~\citep{ramos2003using,palangi2016deep,mitra2017learning} to retrieve relevant documents given a query.
Similarly, we can think of the biomedical entity normalization task as retrieving relevant concepts given a mention~\citep{li2017cnn,mondal2019medical}.
In our work, we use maximum inner product search (MIPS) for retrieving the concepts represented as sparse and dense vectors, whereas previous models could suffer from error propagation of the pipeline approach.

\section{Methodology}

\subsection{Problem Definition}
We define an input mention $m$ as an entity string in a biomedical corpus.
Each input mention has its own CUI $c$ and each CUI has one or more synonyms defined in the dictionary.
The set of synonyms for a CUI is also called as a synset.
We denote the union of all synonyms in a dictionary as $N = [n_1, n_2, \dots]$ where $n \in N$ is a single synonym string.
Our goal is to predict the gold CUI $c^*$ of the input mention $m$ as follows:
\begin{equation}\label{eqn:cui_retrieve}
c^*=\texttt{CUI}(\operatorname{argmax}_{n \in N} P(n | m ; \theta))
\end{equation}
where \texttt{CUI}($\cdot$) returns the CUI of the synonym $n$ and $\theta$ denotes a trainable parameter of our model.

\subsection{Model Description}\label{sec:model}
The overview of our framework is illustrated in Figure~\ref{fig:biosyn}.
We first represent each input mention $m$ and each synonym $n$ in a dictionary using sparse and dense representations.
We treat $m$ and $n$ equally and use a shared encoder for both strings.
During training, we iteratively update top candidates and calculate the marginal probability of the synonyms based on their representations.
At inference time, we find the nearest synonym by performing MIPS over all synonym representations.

\paragraph{Sparse Entity Representation} 
We use tf-idf to obtain a sparse representation of $m$ and $n$.
We denote each sparse representation as $e^s_m$ and $e^s_n$ for the input mention and the synonym, respectively.
tf-idf is calculated based on the character-level n-grams statistics computed over all synonyms $n \in N$. 
We define the sparse scoring function of a mention-synonym pair $(m,n)$ as follows:
\begin{equation}
S_{\text{sparse}}(m,n) = f(e^s_m, e^s_n) \in \mathbb{R}
\end{equation}
\noindent where $f$ denotes a similarity function. We use the inner product between two vectors as $f$.

\paragraph{Dense Entity Representation}
While the sparse representation encodes the morphological information of given strings, the dense representation encodes the semantic information.
Learning effective dense representations is the key challenge in the biomedical entity normalization task~\citep{li2017cnn,mondal2019medical,phan2019robust,fakhraei2019nseen}.
We use pre-trained BioBERT~\citep{10.1093/bioinformatics/btz682} to encode dense representations and fine-tune BioBERT with our synonym marginalization algorithm.\footnote{We used BioBERT v1.1 (+ PubMed) in our work.}
We share the same BioBERT model for encoding mention and synonym representations. 
We compute the dense representation of the mention $m$ as follows:
\begin{equation}
e^d_m = \texttt{BioBERT}(\overline{m})[\texttt{CLS}] \in \mathbb{R}^{h}
\end{equation}
\noindent where $\overline{m}=\{ \overline{m}_1,...,\overline{m}_l \}$ is a sequence of sub-tokens of the mention $m$ segmented by the WordPiece tokenizer~\citep{Wu2016google} and $h$ denotes the hidden dimension of BioBERT (i.e., $h=768$).
[\texttt{CLS}] denotes the special token that BERT-style models use to compute a single representative vector of an input.
The synonym representation $e^d_n \in \mathbb{R}^h$ is computed similarly.
We denote the dense scoring function of a mention-synonym pair $(m,n)$ using the dense representations as follows:
\begin{equation}
S_{\text{dense}}(m,n) = f(e^d_m, e^d_n) \in \mathbb{R}
\end{equation}
where we again used the inner product for $f$.

\paragraph{Similarity Function}
Based on the two similarity functions $S_{\text{sparse}}(m,n)$ and $S_{\text{dense}}(m,n)$, we now define the final similarity function $S(m,n)$ indicating the similarity between an input mention $m$ and a synonym $n$:
\begin{equation}
S(m,n) = S_{\text{dense}}(m,n) + \lambda S_{\text{sparse}}(m,n) \in \mathbb{R}
\end{equation}
\noindent where $\lambda$ is a trainable scalar weight for the sparse score.
Using $\lambda$, our model learns to balance the importance between the sparse similarity and the dense similarity.

\subsection{Training}
The most common way to train the entity representation model is to build a pair-wise training dataset.
While it is relatively convenient to sample positive pairs using synonyms, sampling negative pairs are trickier than sampling positive pairs as there are a vast number of negative candidates.
For instance, the mention `\textit{alpha conotoxin}' (MeSH:D020916) has 6 positive synonyms while its dictionary has 407,247 synonyms each of which can be a negative sampling candidate.
Models trained on these pair-wise datasets often rely on the quality of the negative sampling~\citep{leaman2013dnorm,li2017cnn,phan2019robust,fakhraei2019nseen}.
On the other hand, we use a model-based candidate retrieval and maximize the marginal probability of positive synonyms in the candidates.

\paragraph{Iterative Candidate Retrieval}
Due to a large number of candidates present in the dictionary, we need to retrieve a smaller number of candidates for training.
In our framework, we use our entity encoder to update the top candidates iteratively.
Let $k$ be the number of top candidates to be retrieved for training and $\alpha$~($0 \leq \alpha \leq 1$) be the ratio of candidates retrieved from $S_{\text{dense}}$.
We call $\alpha$ as the dense ratio and $\alpha = 1$ means consisting the candidates with $S_{\text{dense}}$ only.
First, we compute the sparse scores $S_{\text{sparse}}$ and the dense scores $S_{\text{dense}}$ for all $n \in N$.
Then we retrieve the $k - \floor{\alpha k}$ highest candidates using $S_{\text{sparse}}$, which we call as sparse candidates.
Likewise, we retrieve the $\floor{\alpha k}$ highest candidates using $S_{\text{dense}}$, which we call as dense candidates.
Whenever the dense and sparse candidates overlap, we add more dense candidates to match the number of candidates as $k$.
While the sparse candidates for a mention will always be the same as they are based on the static tf-idf representation, the dense candidates change every epoch as our model learns better dense representations.

Our iterative candidate retrieval method has the following benefits.
First, it makes top candidates to have more difficult negative samples as our model is trained, hence helping our model represent a more accurate dense representation of each entity.
Also, it increases the chances of retrieving previously unseen positive samples in the top candidates.
As we will see, comprising the candidates purely with sparse candidates have a strict upper bound while ours with dense candidates can maximize the upper bound.

\paragraph{Synonym Marginalization}
Given the top candidates from iterative candidate retrieval, we maximize the marginal probability of positive synonyms, which we call as synonym marginalization.
Given the top candidates $N_{1:k}$ computed from our model, the probability of each synonym is obtained as:
\begin{equation}
P(n | m;\theta) = \frac{\exp (S(n, m))}{\sum_{n^{\prime} \in N_{1:k}} \exp (S(n^{\prime}, m))}
\end{equation}
\noindent where the summation in the denominator is over the top candidates $N_{1:k}$.
Then, the marginal probability of the positive synonyms of a mention $m$ is defined as follows:
\begin{equation}
P'(m,N_{1:k}) = \sum_{\substack{n \in N_{1:k} \\ \textsc{EQUAL}(m,n)=1 \\ }} P (n | m;\theta)
\end{equation}
\noindent where $\textsc{EQUAL}(m,n)$ is 1 when $\texttt{CUI}(m)$ is equivalent to $\texttt{CUI}(n)$ and 0 otherwise.
Finally, we minimize the negative marginal log-likelihood of synonyms.
We define the loss function of our model as follows:
\begin{equation}
\mathcal{L} = -\frac{1}{M} \sum_{i=1}^M \log P'(m_i,N_{1:k})
\end{equation}
\noindent where $M$ is the number of training mentions in our dataset.
We use mini-batch for the training and use Adam optimizer~\citep{kingma2014adam} to minimize the loss.


\subsection{Inference}
At inference time, we retrieve the nearest synonym of a mention representation using MIPS.
We compute the similarity score $S(m,n)$ between the input mention $m$ and all synonyms $n \in N$ using the inner product and return the CUI of the nearest candidate.
Note that it is computationally cheap to find the nearest neighbors once we pre-compute the dense and sparse representations of all synonyms.

\section{Experimental Setup}

\subsection{Implementation Details}\label{sec:impl}
We perform basic pre-processings such as lower-casing all characters and removing the punctuation for both mentions and synonyms.
To resolve the typo issues in mentions from NCBI disease, we apply the spelling check algorithm following the previous work~\citep{dsouza-ng-2015-sieve}.
Abbreviations of entities are widely used in biomedical entities for an efficient notation which makes the normalization task more challenging. 
Therefore, we use the abbreviation resolution module called Ab3P\footnote{\href{https://github.com/ncbi-nlp/Ab3P}{https://github.com/ncbi-nlp/Ab3P}} to detect the local abbreviations and expand it to its definition from the context~\citep{sohn2008abbreviation}. 
We also split composite mentions (e.g. 'breast and ovarian cancer') into separate mentions (e.g. 'breast cancer' and 'ovarian cancer') using heuristic rules described in the previous work~\citep{dsouza-ng-2015-sieve}.
We also merge mentions in the training set to the dictionary to increase the coverage following the previous work~\citep{dsouza-ng-2015-sieve}.

For sparse representations, we use character-level uni-, bi-grams for tf-idf.
The maximum sequence length of BioBERT is set to 25\footnote{This covers 99.9\% of strings in all datasets.} and any string over the maximum length is truncated to 25.
The number of top candidates $k$ is 20 and the dense ratio $\alpha$ for the candidate retrieval is set to 0.5.
We set the learning rate to 1e-5, weight decay to 1e-2, and the mini-batch size to 16.
We found that the trainable scalar $\lambda$ converges to different values between 2 to 4 on each dataset.
We train \ours~for 10 epochs for NCBI Disease, BC5CDR Disease, and TAC2017 ADR and 5 epochs for BC5CDR Chemical due to its large dictionary size.
Except the number of epochs, we use the same hyperparameters for all datasets and experiments.

We use the top $k$ accuracy as an evaluation metric following the previous works in biomedical entity normalization tasks~\citep{dsouza-ng-2015-sieve,li2017cnn,wright2019normco,phan2019robust,ji2019bert,mondal2019medical}.
We define \text{Acc@}$k$ as 1 if a correct CUI is included in the top $k$ predictions, otherwise 0.
We evaluate our models using \accone~and \accfive.
Note that we treat predictions for composite entities as correct if every prediction for each separate mention is correct.

\begin{table}
\centering
\resizebox{\columnwidth}{!}{
\begin{tabular}{lcccccc}
\toprule
\multirow{2}{*}{Dataset} & \multicolumn{3}{c}{Documents} & \multicolumn{3}{c}{Mentions} \\
& Train & Dev & Test & Train & Dev & Test\\
\midrule
NCBI Disease & \multirow{1}{*}{592} & \multicolumn{1}{c}{\multirow{1}{*}{100}} & \multirow{1}{*}{100} & \multirow{1}{*}{5,134} & \multicolumn{1}{c}{\multirow{1}{*}{787}} & \multirow{1}{*}{960} \\
BC5CDR Disease & \multirow{1}{*}{500} & \multirow{1}{*}{500} & \multirow{1}{*}{500} & \multirow{1}{*}{4,182} & \multirow{1}{*}{4,244} & \multirow{1}{*}{4,424}\\
BC5CDR Chemical & \multirow{1}{*}{500} & \multirow{1}{*}{500} & \multirow{1}{*}{500} & \multirow{1}{*}{5,203} & \multirow{1}{*}{5,347} & \multirow{1}{*}{5,385}\\
TAC2017ADR & \multirow{1}{*}{101} & \multicolumn{1}{c}{\multirow{1}{*}{-}} & \multirow{1}{*}{99} & \multirow{1}{*}{7,038} & \multicolumn{1}{c}{\multirow{1}{*}{-}} & \multirow{1}{*}{6,343}\\ 
\bottomrule
\end{tabular}
}
\caption{Data statistics of four biomedical entity normalization datasets. See Section~\ref{sec:dataset} for more details of each dataset.}\label{table:data_staticstics}
\vspace{-.5cm}
\end{table}

\begin{table*}[ht!]
\centering
\begin{threeparttable}
\resizebox{0.95\textwidth}{!}{
\begin{tabular}{l *{4}{c c}}
\toprule
\multirow{2}{*}{Models} & \multicolumn{2}{c}{NCBI Disease} & \multicolumn{2}{c}{BC5CDR Disease} & \multicolumn{2}{c}{BC5CDR Chemical} & \multicolumn{2}{c}{TAC2017ADR} \\
 & \accone & \accfive & \accone & \accfive & \accone & \accfive & \accone & \accfive \\ 
\midrule
Sieve-Based~\citep{dsouza-ng-2015-sieve}    & 84.7 &  -  & 84.1 & - &  90.7$^\dagger$   & - &  84.3$^\dagger$  & - \\
Taggerone~\citep{leaman2016taggerone} &  87.7   &  -  & 88.9 & - & 94.1 & - &  -   & - \\
CNN Ranking~\citep{li2017cnn}         & 86.1 &  -  &  -   & - &  -   & - &  -   & - \\
NormCo~\citep{wright2019normco}       &  87.8   &  -  & 88.0 & - &  -   & - &  -   & - \\
BNE~\citep{phan2019robust}            &  87.7   &  -  & 90.6 & - & 95.8 & - &  -   & - \\
BERT Ranking~\citep{ji2019bert}       & 89.1 &  -  &  -   & - &  -   & - & 93.2 & - \\
TripletNet~\citep{mondal2019medical}  & 90.0 &  -  &  -   & - &  -   & - &  -   & - \\
\midrule
\ours~(\textsc{s-score})    & 87.6 & 90.5 & 92.4 & 95.7 & 95.9 & 96.8 & 91.4 & 94.5 \\
\ours~(\textsc{d-score})     & 90.7 & 93.5 & 92.9 & 96.5 & \textbf{96.6} & 97.2 & 95.5 & 97.5 \\
\ours~($\alpha=0.0$)         & 89.9 & 93.3 & 92.2 & 94.9 &  96.3 &  97.2 &   95.3   & 97.6 \\
\ours~($\alpha=1.0$)          & 90.5 & \textbf{94.5} & 92.8 & \textbf{96.0} &  96.4   &  \textbf{97.3}   &   \textbf{95.8}   & \textbf{97.9} \\
\midrule
\ours~(Ours)        & \textbf{91.1} & 93.9 & \textbf{93.2} & \textbf{96.0} & \textbf{96.6} & 97.2 & 95.6 & 97.5 \\
\bottomrule
\end{tabular}
}
\begin{tablenotes}
\item[$\dagger$] {\footnotesize We used the author's provided implementation to evaluate the model on these datasets.}
\end{tablenotes}
\end{threeparttable}
\caption{Experimental results on four biomedical entity normalization datasets
}
\label{table:bc5di_result}
\end{table*}

\begin{figure*}[!t]
\centering
\subfloat[NCBI Disease]{\includegraphics[width=5.2cm,height=3.2cm]{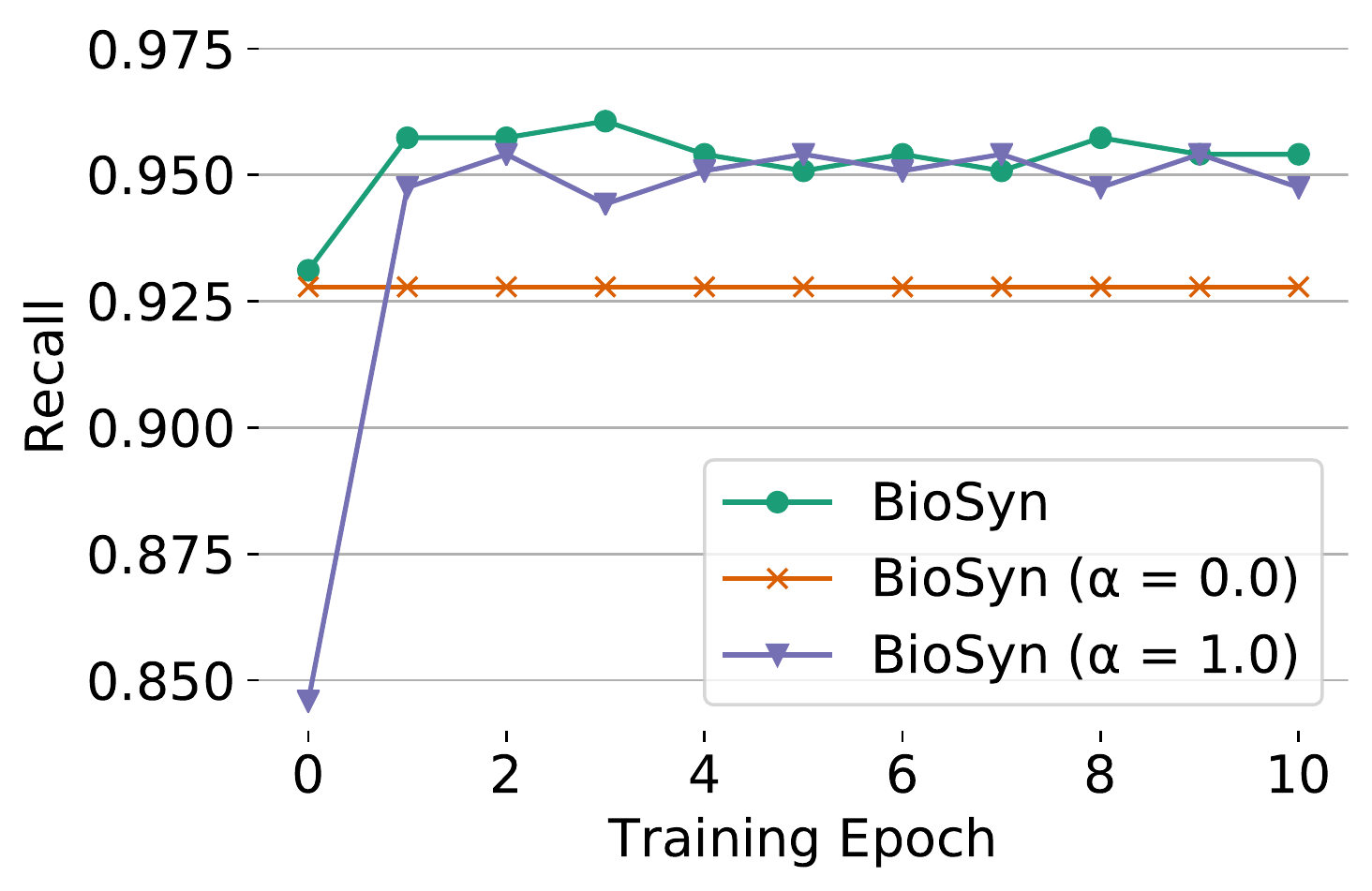}}
\subfloat[BC5CDR Disease]{\includegraphics[width=5.2cm,height=3.2cm]{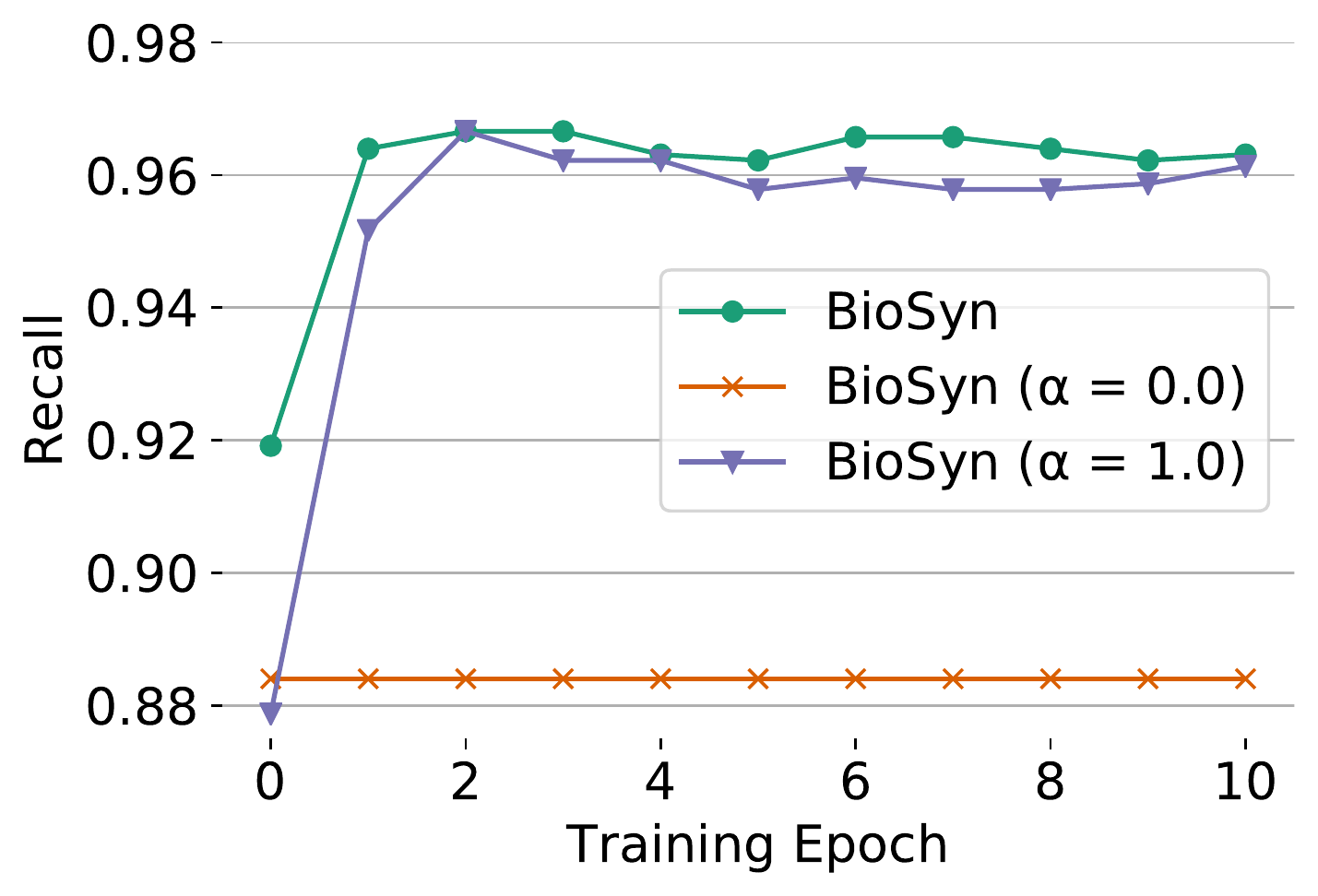}}
\subfloat[BC5CDR Chemical]{\includegraphics[width=5.2cm,height=3.2cm]{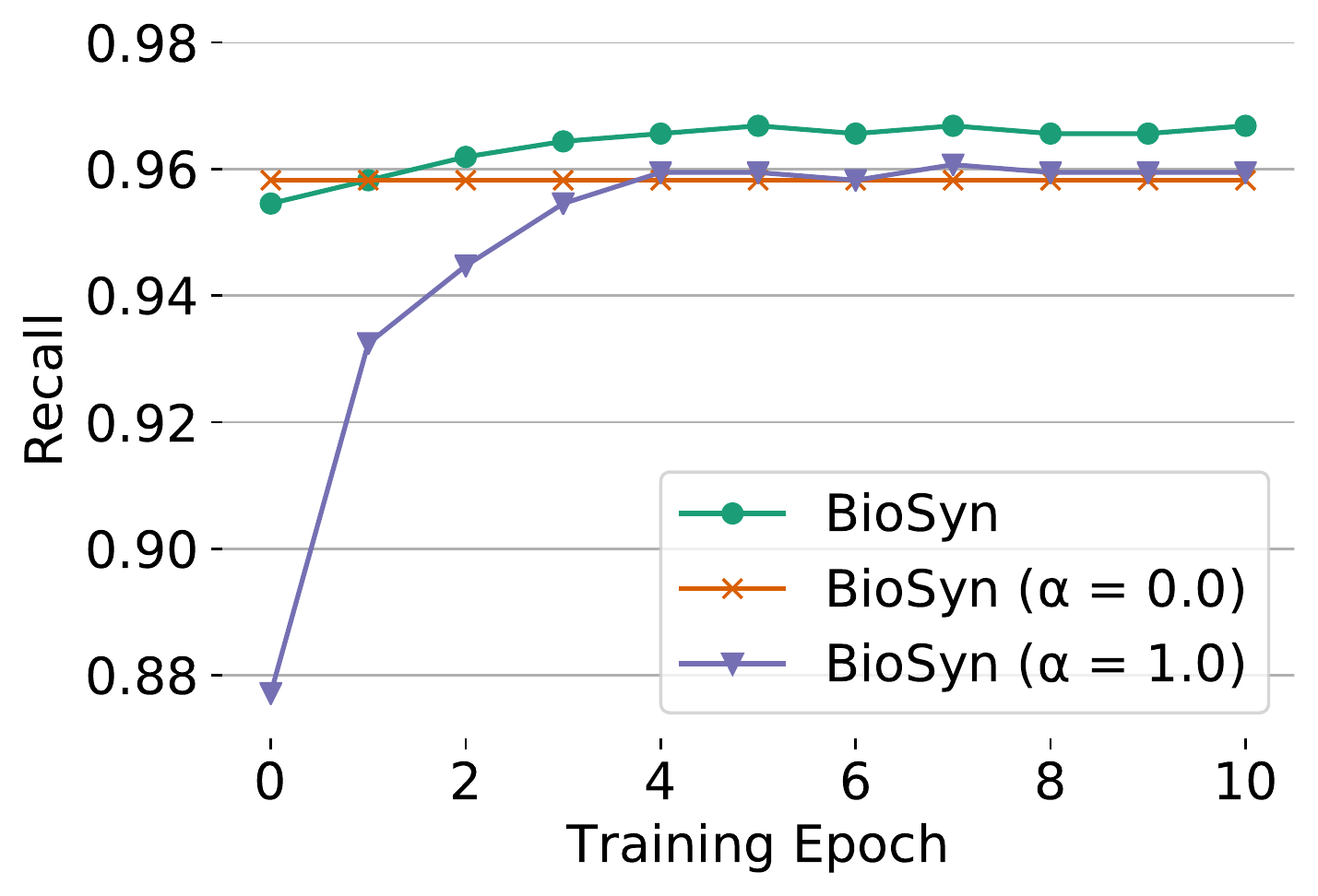}}
\caption{Effect of iterative candidate retrieval on the development sets of NCBI Disease, BC5CDR Disease, and BC5CDR Chemical. We show the recall of top candidates of each model.
}\label{figure:topk_accuracy}
\end{figure*}

\subsection{Datasets}\label{sec:dataset}
We use four biomedical entity normalization datasets having three different biomedical entity types (disease, chemical, adverse reaction).
The statistics of each dataset is described in Table~\ref{table:data_staticstics}.

\paragraph{NCBI Disease Corpus}
NCBI Disease Corpus~\citep{dougan2014ncbi}\footnote{\href{https://www.ncbi.nlm.nih.gov/CBBresearch/Dogan/DISEASE}{https://www.ncbi.nlm.nih.gov/CBBresearch/Dogan/DISEASE}} provides manually annotated disease mentions in each document with each CUI mapped into the MEDIC dictionary~\citep{davis2012medic}.
In this work, we use the July 6, 2012 version of MEDIC containing 11,915 CUIs and 71,923 synonyms included in MeSH and/or OMIM ontologies.

\paragraph{Biocreative V CDR}

BioCreative V CDR~\citep{li2016biocreative}\footnote{\href{https://biocreative.bioinformatics.udel.edu/tasks/biocreative-v/track-3-cdr}{https://biocreative.bioinformatics.udel.edu/tasks/biocreative-v/track-3-cdr}} is a challenge for the tasks of chemical-induced disease (CID) relation extraction.
It provides disease and chemical type entities.
The annotated disease mentions in the dataset are mapped into the MEDIC dictionary like the NCBI disease corpus.
The annotated chemical mentions in the dataset are mapped into the Comparative Toxicogenomics Database (CTD)~\citep{davis2018comparative} chemical dictionary.
In this work, we use the November 4, 2019 version of the CTD chemical dictionary containing 171,203 CUIs and 407,247 synonyms included in MeSH ontologies.
Following the previous work~\citep{phan2019robust}, we filter out mentions whose CUIs do not exist in the dictionary.

\paragraph{TAC2017ADR}
TAC2017ADR~\citep{roberts2017overview}\footnote{\href{https://bionlp.nlm.nih.gov/tac2017adversereactions}{https://bionlp.nlm.nih.gov/tac2017adversereactions}} is a challenge whose purpose of the task is to extract information on adverse reactions found in structured product labels.
It provides manually annotated mentions of adverse reactions that are mapped into the MedDRA dictionary~\citep{brown1999medical}. 
In this work, we use MedDRA v18.1 which contains 23,668 CUIs and 76,817 synonyms. 

\section{Experimental Results}
We use five different versions of our model to see the effect of each module in our framework.
First, \ours~denotes our proposed model with default hyperparameters described in Section~\ref{sec:impl}.
\ours~$(\textsc{s-score})$ and \ours~$(\textsc{d-score})$ use only sparse scores or dense scores for the predictions at inference time, respectively.
To see the effect of different dense ratios, \ours~($\alpha=0.0$) uses only sparse candidates and \ours~($\alpha=1.0$) uses only dense candidates during training.

\subsection{Main Results}
Table~\ref{table:bc5di_result} shows our main results on the four datasets.
Our model outperforms all previous models on the four datasets and achieves new state-of-the-art performance.
The \accone~improvement on NCBI Disease, BC5CDR Disease, BC5CDR Chemical and TAC2017ADR are 1.1\%, 2.6\%, 0.8\% and 2.4\%, respectively.
Training with only dense candidates ($\alpha=1.0$) often achieves higher \accfive~than \ours~showing the effectiveness of dense candidates.

\subsection{Effect of Iterative Candidate Retrieval}
In Figure~\ref{figure:topk_accuracy}, we show the effect of the iterative candidate retrieval method.
We plot the recall of top candidates used in each model on the development sets.
The recall is 1 if any top candidate has the gold CUI.
\ours~($\alpha=1$) uses only dense candidates while \ours~($\alpha=0$) uses sparse candidates.
\ours~utilizes both dense and sparse candidates.
Compared to the fixed recall of \ours~($\alpha=0$), we observe a consistent improvement in \ours~($\alpha=1$) and \ours.
This proves that our proposed model can increase the upper bound of candidate retrieval using dense representations.

\subsection{Effect of the Number of Candidates}
\begin{figure}[!t]
\centering
\includegraphics[width=0.9\columnwidth]{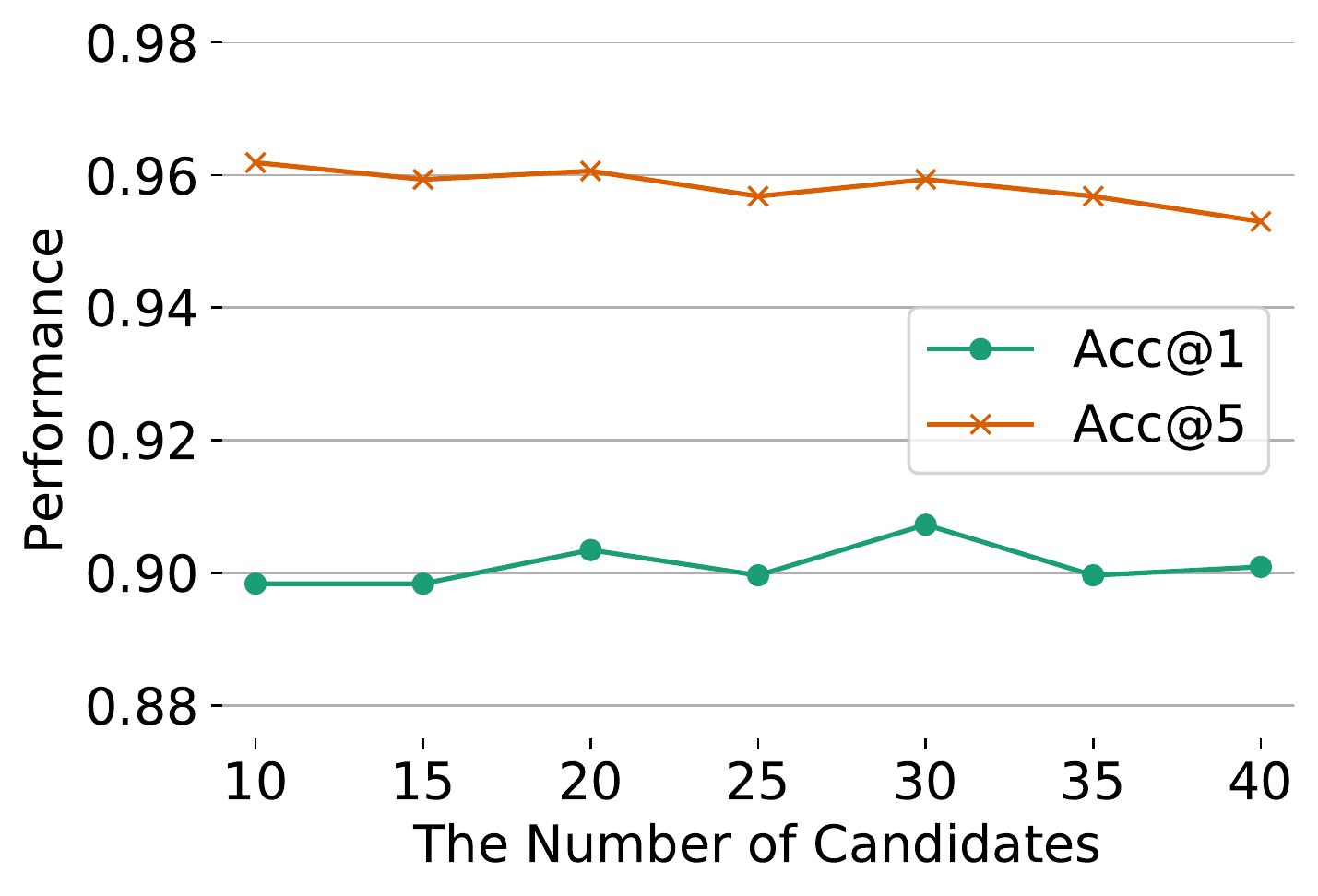}
\caption{Performance of \ours~on the development set of NCBI Disease with different numbers of candidates
}
\label{figure:topk}
\end{figure}

We perform experiments by varying the number of top candidates used for training.
Figure~\ref{figure:topk} shows that a model with 20 candidates performs reasonably well in terms of both~\accone ~and~\accfive.
It shows that more candidates do not guarantee higher performance, and considering the training complexity, we choose $k=20$ for all experiments.

\subsection{Effect of Synonym Marginalization}

Our synonym marginalization method uses marginal maximum likelihood (MML) as the objective function. 
To verify the effectiveness of our proposed method, we compare our method with two different strategies: hard EM~\citep{liang2018memory} and the standard pair-wise training~\citep{leaman2013dnorm}.
The difference between hard EM and MML is that hard EM maximizes the probability of a single positive candidate having the highest probability.
In contrast, MML maximizes marginalized probabilities of all synonyms in the top candidates.
For hard EM, we first obtain a target $\tilde{n}$ as follows:
\begin{equation}
\tilde{n}=\operatorname{argmax}_{n \in N_{1:k}} P(n | m ; \theta)
\end{equation}
\noindent where most notations are the same as Equation~\ref{eqn:cui_retrieve}.
The loss function of hard EM is computed as follows:
\begin{equation}
\mathcal{L} = -\frac{1}{M} \sum_{i=1}^{M} \log P(\tilde{n} | m_i;\theta).
\end{equation}

The pair-wise training requires a binary classification model.
For the pair-wise training, we minimize the binary cross-entropy loss using samples created by pairing each positive and negative candidate in the top candidates with the input mention.
Table~\ref{table:synonym_marginalization} shows the results of applying three different loss functions on BC5CDR Disease and BC5CDR Chemical.
The results show that MML used in our framework learns better semantic representations than other methods.


\begin{table}[t]
\centering
\resizebox{\columnwidth}{!}{
\begin{tabular}{lcccc}
\toprule
\multirow{2}{*}{Methods} & 
\multicolumn{2}{c}{BC5CDR D.}&
\multicolumn{2}{c}{BC5CDR C.}\\ 
& \accone & \accfive & \accone & \accfive \\ \midrule
MML & \textbf{91.1} & 95.4 & \textbf{96.7} & \textbf{97.7} \\ 
Hard EM & 91.0 & \textbf{95.8} & 96.5 & 97.5 \\ 
Pair-Wise Training & 90.7 & 94.4 & 96.3 & 97.2 \\ 
\bottomrule
\end{tabular}
}
\caption{\label{table:synonym_marginalization} Comparison of two different training methods on the development sets of BC5CDR Disease, BC5CDR Chemical
}
\end{table}

\section{Analysis}

\subsection{Iterative Candidate Samples}

\begin{table*}[ht]
    \centering
    \resizebox{\textwidth}{!}{
        \begin{tabular}{c| c c c c}
        \toprule
         Rank & tf-idf & Epoch 0 & Epoch 1 & Epoch 5 \\
        \midrule
        \multicolumn{5}{c}{prostate carcinomas (MeSH:D011471)}
        \\\midrule
1 &  carcinomas               & \textbf{prostatic cancers*}&  \textbf{prostate cancers*}       &  \textbf{prostate cancers*}    \\
 2 &  teratocarcinomas         &  \textbf{prostate cancers*}   &  \textbf{prostatic cancers*}        &  \textbf{prostate cancer*}  \\
  3 &  pancreatic carcinomas ...&  glioblastomas                &  \textbf{prostate neoplasms*}       &  \textbf{prostatic cancers*} \\
 4 &  carcinomatoses           &  carcinomas                   &  \textbf{prostate cancer*}          &  \textbf{prostate neoplasms*} \\
 5 &  carcinomatosis           &  renal cell cancers           &  \textbf{prostate neoplasm*}        &  \textbf{prostatic cancer*}  \\
 6 &  breast carcinomas        &  renal cancers                &  \textbf{prostatic cancer*}         &  \textbf{cancers prostate*}  \\
 7 &  teratocarcinoma          &  retinoblastomas              &  \textbf{prostatic neoplasms*}      &  \textbf{prostate neoplasm*} \\
 8 &  carcinoma                &  cholangiocarcinomas          &  \textbf{advanced prostate cancers*}&  \textbf{cancer of prostate*} \\
 9 &  breast carcinoma         &  pulmonary cancers            &  \textbf{prostatic neoplasm*}       &  \textbf{cancer of the prostate*} \\
 10&  carcinosarcomas          &  gonadoblastomas              &  prostatic adenomas                 &  \textbf{cancer prostate*} \\  
\midrule
\multicolumn{5}{c}{brain abnormalities (MeSH:D001927)}
\\\midrule
  1 & nail abnormalities                & brain dysfunction minimal      & \textbf{brain pathology*}      & \textbf{brain disorders*}         \\
2 & abnormalities nail                & \textbf{brain pathology*}      & \textbf{brain disorders*}      & \textbf{brain disorder*}          \\
3 & facial abnormalities              & deficits memory                & white matter abnormalities     & \textbf{brain diseases*}          \\
4 & torsion abnormalities             & memory deficits                & \textbf{brain disease*}        & \textbf{brain disease*}           \\
5 & spinal abnormalities              & neurobehavioral manifestations & \textbf{brain diseases*}       & abnormalities in brain dev...\\
6 & skin abnormalities                & white matter diseases          & \textbf{brain disorder*}       & nervous system abnormalities      \\
7 & genital abnormalities             & brain disease metabolic        & neuropathological abnormalities& white matter abnormalities        \\
8 & nail abnormality                  & neuropathological abnormalities& brain dysfunction minimal      & metabolic brain disorders         \\
9 & clinical abnormalities            & neurobehavioral manifestation  & white matter lesions           & brain metabolic disorders         \\
10& abnormalities in brain dev...& \textbf{brain disease*}        & brain injuries                 & \textbf{brain pathology*}        
                \\
\midrule
\multicolumn{5}{c}{type ii deficiency (OMIM:217000)}
\\\midrule
 1 &  mat i iii deficiency             &  deficiency disease             &  \textbf{type ii c2 deficient*}  &  factor ii deficiency            \\
 2 &  naga deficiency type iii ...     &  type 1 citrullinemia           &  deficiency disease              &  \textbf{type ii c2 deficient*}  \\
 3 &  properdin deficiency type iii ...&  cmo ii deficiency              &  deficiency diseases             &  factor ii deficiencies          \\
 4 &  properdin deficiency type i ...  &  mitochondrial trifunctional ...&  \textbf{type ii c2d deficiency*}&  \textbf{type ii c2d deficiency*}\\
 5 &  naga deficiency type iii         &  \textbf{type ii c2 deficient*} &  factor ii deficiency            &  diabetes mellitus type ii       \\
 6 &  naga deficiency type ii          &  deficiency aga                 &  deficiency protein              &  deficiency factor ii            \\
 7 &  properdin deficiency type iii    &  sodium channel myotonia        &  deficiency vitamin              &  \textbf{c2 deficiency*}         \\
 8 &  properdin deficiency type ii     &  deficiency diseases            &  deficiency factor ii            &  t2 deficiency                   \\
 9 &  tc ii deficiency                 &  tuftsin deficiency             &  deficiency arsa                 &  tc ii deficiency                \\
 10&  si deficiency                    &  triosephosphate isomerase ...  &  class ii angle                  &  mitochondrial complex ii ...    \\
\bottomrule

        \end{tabular}
    }
    \caption{Changes in the top 10 candidates given the two input mentions from the NCBI Disease development set. Synonyms having correct CUIs are indicated in boldface with an asterisk.
    } 
    \label{table:qualitative_analysis}
\end{table*}

\begin{table*}[ht!]
    \centering
    \resizebox{0.95\textwidth}{!}{
        \begin{tabular}{lccc|c}
        \toprule
        Error Type & Input & Predicted & Annotated & Statistics \\
        \midrule
        Incomplete Synset & hypomania & hypodermyiasis & mood disorders & 25 (29.4\%) \\
        Contextual Entity & colorectal adenomas & colorectal adenomas & polyps adenomatous & 3 (3.5\%) \\
        Overlapped Entity & desmoid tumors & desmoid tumor & desmoids & 11 (12.9\%) \\
        Abbreviation & sca1 & oca1 & spinocerebellar ataxia 1 & 10 (11.8\%) \\
        Hypernym & campomelia & campomelic syndrome & campomelia cumming type & 10 (11.8\%) \\
        Hyponym & eye movement abnormalities & eye movement disorder & eye abnormalities & 23 (27.1\%) \\
        Others & hamartoma syndromes & hamartomas & multiple hamartoma syndromes & 3 (3.5\%) \\
        \bottomrule
        \end{tabular}
    }
    \caption{Examples and statistics of error cases on the NCBI Disease test set
    }
    \label{table:error_analysis}
\end{table*}

In Table~\ref{table:qualitative_analysis}, we list top candidates of \ours~from the NCBI Disease development set.
Although the initial candidates did not have positive samples due to the limitation of sparse representations, candidates at epoch 1 begin to include more positive candidates.
Candidates at epoch 5 include many positive samples, while negative samples are also closely related to each mention.

\subsection{Error Analysis}

In Table~\ref{table:error_analysis}, we analyze the error cases of our model on the test set of NCBI Disease.
We manually inspected all failure cases and defined the following error cases in the biomedical entity normalization task: Incomplete Synset, Contextual Entity, Overlapped Entity, Abbreviation, Hypernym, and Hyponym.
Remaining failures that are difficult to categorize are grouped as Others. 

\textbf{Incomplete Synset} is the case when the surface form of an input mention is very different from the provided synonyms of a gold CUI and requires the external knowledge for the normalization.
\textbf{Contextual Entity} denotes an error case where an input mention and the predicted synonym are exactly the same but have different CUIs.
This type of error could be due to an annotation error or happen when the same mention can be interpreted differently depending on its context.
\textbf{Overlapped Entity} is an error where there is an overlap between the words of input mention and the predicted candidate.
This includes nested entities.
\textbf{Abbrevation} is an error where an input mention is in an abbreviated form but the resolution has failed even with the external module Ab3P.
\textbf{Hypernym} and \textbf{Hyponym} are the cases when an input mention is a hypernym or a hyponym of the annotated entity.

Based on our analyses, errors are mostly due to ambiguous annotations (Contextual Entity, Overlapped Entity, Hypernym, Hyponym) or failure of pre-processings (Abbreviation).
Incomplete Synset can be resolved with a better dictionary having richer synonym sets.
Given the limitations in annotations, we conclude that the performance of \ours~has almost reached the upper bound.

\section{Conclusion}
In this study, we introduce \ours~that utilizes the synonym marginalization technique and the iterative candidate retrieval for learning biomedical entity representations.
On four biomedical entity normalization datasets, our experiment shows that our model achieves state-of-the-art performance on all datasets, improving previous scores up to \upperb\%.
Although the datasets used in our experiments are in English, we expect that our methodology would work in any language as long as there is a synonym dictionary for the language.
For future work, an extrinsic evaluation of our methods is needed to prove the effectiveness of learned biomedical entity representations and to prove the quality of the entity normalization in downstream tasks.

\section*{Acknowledgments}

This research was supported by National Research Foundation of Korea (NRF-2016M3A9A7916996, NRF-2014M3C9A3063541).  We thank the members of Korea University, and the anonymous reviewers for their insightful comments.

\bibliography{acl2020}
\bibliographystyle{acl_natbib}
\end{document}